\documentclass[a4paper, 10pt, conference]{ieeeconf}      

\IEEEoverridecommandlockouts                              

\usepackage[T1]{fontenc}
\usepackage[ansinew]{inputenc}
 \usepackage{multirow}
\usepackage{adjustbox,lipsum}
\usepackage{lmodern} 
\usepackage{amsfonts}
\usepackage[noend]{algpseudocode}
\usepackage{savesym}
\savesymbol{AND}
\usepackage{algorithm}
\usepackage{graphicx}
\usepackage{amsmath}
\usepackage{mathtools}

\title{\LARGE \bf
Extending the Multiple Traveling Salesman Problem for Scheduling a Fleet of Drones Performing Monitoring Missions
}

\author{Emmanouil S. Rigas, Panayiotis Kolios, Georgios Ellinas
\thanks{The authors are with the Department of Electrical and Computer Engineering and the KIOS Research and Innovation Center of Excellence (KIOS CoE), University of Cyprus, emails: \{rigas.emmanouil, kolios.panayiotis, gellinas\}@ucy.ac.cy}
}

\begin{document}

\maketitle
\thispagestyle{empty}
\pagestyle{empty}

\begin{abstract}
In this paper we schedule the travel path of a set of drones across a graph where the nodes need to be visited multiple times at pre-defined points in time. This is an extension of the well-known multiple traveling salesman problem. The proposed formulation can be applied in several domains such as the monitoring of traffic flows in a transportation network, or the monitoring of remote locations to assist search and rescue missions. Aiming to find the optimal schedule, the problem is formulated as an Integer Linear Program (ILP). Given that the problem is highly combinatorial, the optimal solution scales only for small sized problems. Thus, a greedy algorithm is also proposed that uses a one-step look ahead heuristic search mechanism. In a detailed evaluation, it is observed that the greedy algorithm has near-optimal performance as it is on average at $92.06\%$ of the optimal, while it can potentially scale up to settings with hundreds of drones and locations.    
\end{abstract}

\section{Introduction}
\label{sec:intro}

Unmanned aerial vehicles (UAVs), or simply drones \cite{villasenor2012drone, kyrkou2019drones}, are used in a plethora of civil applications due to their ease of deployment, low maintenance cost, high-mobility and ability to hover \cite{8682048}. Such vehicles are utilized for real time monitoring of road traffic, search and rescue operations, civil infrastructure inspection, wireless coverage, remote sensing, delivery of goods, security and surveillance, and precision agriculture \cite{floreano2015science}.

A main advantage of drones is that, in contrast to other vehicles, they are not restricted to traveling over a road network and thus can swiftly move over disperse locations. In order to maximize this ability, their scheduled and coordinated flying is crucial. Thus, in this paper we propose a generalized graph-based solution to schedule drones for monitoring missions. Here, it is important to note that both surveillance and monitoring tasks focus on developing control laws which enable groups of robots to transverse and observe a given domain, but with a slightly different focus. The goal of surveillance is to maximize some measure of coverage or information gathering, while monitoring focuses on ensuring that certain areas of the domain (usually predefined) are visited with a certain frequency.

Recently, drone-related problems have been intensively researched and artificial intelligence (AI) techniques such as heuristic search, optimization, multi-agent systems and machine learning are becoming extremely popular. For example, Kitjacharoenchai et al. \cite{kitjacharoenchai2019multiple} study a delivery truck-drone combination problem, where autonomous drones fly from delivery trucks, make deliveries, and subsequently fly to any available delivery truck nearby. Aiming to minimize the arrival time of both trucks and drones at the depot after completing the deliveries, they propose two solutions, one using mixed integer programming and another using insertion heuristics to solve large sized problems. Moreover, Evers et al. \cite{evers2014online} study online stochastic UAV mission planning with time windows and time-sensitive targets using heuristic methods. In addition, Delle Fave et al. \cite{delle2012deploying} study the use of drones for aerial imagery collection and they define this problem as one of task assignment where the drones dynamically coordinate over tasks representing the imagery collection requests. They solve the resulting optimization problem using an asynchronous and decentralized implementation of the max-sum algorithm. In a slightly different vein, Ramchurn et al. \cite{ramchurn2016disaster} model the drones as intelligent agents and study the way human-agent collectives can address challenges in disaster response. Specifically, their proposed methodology utilizes crowd-sourcing combined with machine learning to obtain situational awareness from large streams of reports posted by members of the public. This collected information can be utilized to inform human-agent teams in coordinating multi-UAV deployments, as well as task planning for responders on the ground. In a similar vein, Baker et al. \cite{baker2016planning} study the survivor discovery problem and present a solution based on a continuous factored coordinated Monte Carlo tree search algorithm. Further, Sharafeddine and Islambouli \cite{sharafeddine2019demand} study the use of UAVs as instant recovery devices for cellular networks aiming to decide on the initial location of the UAVs. The authors propose an optimal solution using Mixed Integer Linear Programming techniques as well as an equivalent greedy algorithm. Finally, Menouar et al. \cite{7876852} outline the possible applications of drones for supporting intelligent transportation systems within a smart city domain. 

Against this background, we initially formulate the problem of scheduling drones across a set of locations with monitoring demand in predefined points in time as an Integer Linear Program (ILP) and we solve it offline and optimally. Given the high complexity of the problem and the equivalent limited scalability of the optimal solution, we also develop a greedy algorithm that uses heuristic search. In our experimental evaluation, we observe that the greedy algorithm has very good scalability and performance close to the optimal. This problem is an extension of the Multiple Traveling Salesman Problem (MTSP) \cite{bektas2006multiple}, which is based on the well-known Traveling Salesman Problem (TSP) \cite{miller1960integer}. Compared to the MTSP, in this work (i) all agents do not begin their journey from the same node, and (ii) an agent may pass from one node multiple times in predefined points in time. This problem also shares similarities with the Vehicle Routing Problem \cite{Dantzig:1959:TDP:2780402.2780408} but differs in point (ii) mentioned above. The model of this problem is based on \cite{RIGAS2018248} where the authors study the scheduling of electric vehicles in a mobility-on-demand scheme. The main differences are that (a) in this work the agents initiate their traveling in an autonomous manner, and (b) the execution of a trip is not directly related to the execution of the previous one.   

The rest of the paper is structured as follows: Section~\ref{sec:proDef} provides a detailed problem formulation, Section~\ref{sec:opt} describes the optimal solution of the problem and Section~\ref{sec:greedy} the equivalent greedy one. Section~\ref{sec:eval} evaluates the proposed solutions in different settings and finally, Section~\ref{sec:con} concludes this work and provides insights for future work.

\section{Problem definition}
\label{sec:proDef}

We define the set of UAVs or drones $a \in A \subseteq \mathbb{N}$ acting as fully cooperative agents. Each agent has its own type defined by a tuple $p_{a}=\{n_{a}^{init}, n_{a}^{fin}, n_{a, t}, e_{a, t}, e_{a}^{max}, v_{a, t}, v_{a}^{max}\}$ where $n_{a}^{init}, n_{a}^{fin}, n_{a, t}$ are the initial, final and current location of the agent, $e_{a, t}, e_{a}^{max}$ are the current and maximum energy level of the agent, $v_{a, t}$ is the current velocity and $v_{a}^{max}$ the maximum velocity of the agent. We consider a set of discrete points in time $T \subset \mathbb{N}$, $t \in T$ where time is global for the system and the same for all agents. 

All agents are supposed to move across an undirected fully connected graph $G(N,E)$ where $n \in N \subseteq \mathbb{N}$ is a set of nodes and $\{i,j\} \in E \subseteq \mathbb{N}$ is a set of edges. Every edge has a cost $c_{\{i,j\}} \in \mathbb{R}$ that denotes the time to travel across an edge, or the equivalent energy required. 

The system aims to schedule the agents to pass from specific nodes of the graph at specific points in time. To achieve this, we define the ``flying demand'' as $d_{n,t} \in \{0, 1\}$ which contains the set of points in time a node $n$ should be visited by an agent. The trips of all agents over the edges of the graph must be scheduled and coordinated in order for the flying demand of each node to be covered to the maximum extend (i.e., fully covering the demand may be impossible due to insufficient resources). In this setting, we assume that a fully charged battery is enough for a drone to cover all trips concerned and that the agents will not collide while flying over the same edge as they fly in different altitudes. In this work, henceforth, the terms \textit{agent}, \textit{UAV} and \textit{drone} are used interchangeably.

\section{Optimal solution}
\label{sec:opt}

In this section, we model the problem of scheduling drones as an Integer Linear Program (ILP) and we solve it optimally using IBM Ilog CPLEX $12.10$. We define $4$ decision variables: (1) $fd_{n,t} \in \{0,1\}$ which denotes whether at least one agent flies over node $n$ at time $t$, (2) $ep_{n,a,t} \in \{0,1\}$ which denotes whether agent $a$ hovers over location $n$ at time $t$, (3) $k_{n,n',a,t} \in \{0,1\}$ which denotes whether agent $a$ flies across the edge connecting nodes $n$ and $n': n'\neq n$ at time $t$ and (4) $startT_{n,n',a,t} \in \{0,1\}$ which denotes the time $t$ an agent $a$ begins traveling across the edge connecting nodes $n$ and $n': n'\neq n$.

We define an objective function (Eq.~\ref{eq:obj}) which maximizes the demand that is actually covered by the agents. This function consists of the sum of the points in time an agent passed by a node based on the initial demand and the sum of all agents' location changes multiplied by a very small number $\mu$. The second sum is always smaller than the first and is subtracted from it, in order to prevent agents from changing locations when they do not need to do so. The reader should note that this function is linearized at run time by CPLEX. This is usually done by adding two extra decision variables and two extra constraints. The same is true for all absolute values used later in the constraints. This objective function is maximized under a number of constraints:

\textbf{Objective function:} 

{\small
\begin{equation}
\begin{split}
\label{eq:obj}
    \sum_{n \in N}\sum_{t \in T}(d_{n,t} \times fd_{n,t}) - \\ (\sum_{n \in N}\sum_{n' \in N}\sum_{a \in A}\sum_{t \in T-1}(|k_{n,n',a,t+1} - k_{n,n',a,t}|)) \times \mu
\end{split}
\end{equation}
}

\textbf{Subject to:}

\textit{Temporal, spatial, and routing constraints:}

{\small
\begin{equation}
\label{eq:con1}
    ep_{n_a^{init},a,t=0}= 1, \forall a \in A
\end{equation}

\begin{equation}
\label{eq:con2}
    ep_{n_a^{fin},a,t=T}= 1, \forall a \in A
\end{equation}

\begin{equation}
\label{eq:con3}
    \sum_{n \in N}ep_{n,a,t} \leq 1, \forall a \in A, t \in T
\end{equation}

\begin{equation}
\label{eq:con4}
    \sum_{n \in N}\sum_{n' \in N}k_{n,n',a,t} \leq 1, \forall a \in A, t \in T
\end{equation}

\begin{equation}
\label{eq:con5}
    \sum_{n \in N}\sum_{n' \in N}k_{n,n',a,t} = 1- \sum_{n \in N}ep_{n,a,t}, \forall a \in A, t \in T
\end{equation}

\begin{equation}
\label{eq:con7}
    \sum_{n \in N}\sum_{n' \in N}startT_{n,n',a,t} \leq 1, \forall a \in A, t \in T
\end{equation}

\begin{equation}
\begin{split}
\label{eq:con9}
    ep_{n,a,t} - ep_{n,a,t+1} \leq \sum_{n' \in N} k_{n,n',a,t}, \\ \forall n \in N, t: t \geq 0 \& t\leq T-1, a \in A
\end{split}
\end{equation}

\begin{equation}
\begin{split}
\label{eq:con10}
    ep_{n,a,t} - ep_{n,a,t-1} \leq \sum_{n' \in N} k_{n,n',a,t}, \\ \forall n \in N, t: t \geq 1 \& t\leq T, a \in A
\end{split}
\end{equation}

\begin{equation}
\begin{split}
\label{eq:con12}
    \sum_{t \in T}|startT_{n,n',a,t+1}-startT_{n,n',a,t}| = \\ \sum_{t \in T}|k{n,n',a,t+1}-k_{n,n',a,t}|, \forall n \in N, n \in N, a \in A
\end{split}
\end{equation}

\begin{equation}
\begin{split}
\label{eq:con13}
    \sum_{n \in N}\sum_{t \in T-1}\sum_{a \in A} |ep_{n,a,t+1}-ep_{n,a,t}| = \\ \sum_{n \in N}\sum_{n' \in N}\sum_{t \in T-1}\sum_{a \in A} |startT_{n,n',a,t+1}-startT_{n,n',a,t}|
\end{split}
\end{equation}
}
\textit{Completion constraints:}
{\small
\begin{equation}
\label{eq:con6}
    fd_{n,t} \leq \sum_{a \in A} ep_{n,a,t}, \forall n \in N, t \in T
\end{equation}

\begin{equation}
\begin{split}
\label{eq:con11}
    ep_{n,a,t} - ep_{n,a,t+1} \leq \sum_{n' \in N} startT_{n,n',a,t},\\ \forall n \in N, t: t \geq 0 \mbox{ \& } t\leq T-1, a \in A
\end{split}
\end{equation}

\begin{equation}
\begin{split}
\label{eq:con8}
    \sum_{t \in T}\sum_{n \in N}\sum_{n' \in N} k_{n,n',a,t} =\\ \sum_{t \in T}\sum_{n \in N}\sum_{n' \in N} startT_{n,n',a,t} \times c_{\{n,n'\}}, \forall a \in A
\end{split}
\end{equation}

\begin{equation}
\begin{split}
\label{eq:con14}
    \sum_{t':t'\geq t \& t' \leq t+c_{\{n,n'\}}+1 \mbox{ \& } t':t+c_{\{n,n'\}}+1<T}k_{n,n',a,t'} \geq \\ startT_{n,n',a,t} \times c_{\{n,n'\}}, \forall n \in N, n' \in N, t \in T, a \in A 
\end{split}
\end{equation}
}

The \textit{temporal, spatial, and routing constraints} ensure the proper placement of the drones over time. Thus, each agent must be at its initial location at time $t=0$ (Eq.~\ref{eq:con1}) and at its final location at time $t=T$ (Eq.~\ref{eq:con2}). Moreover, each agent can fly over at most one location in each point in time (Eq.~\ref{eq:con3}) and it can travel across at most one edge in each point in time (Eq.~\ref{eq:con4}). In addition, each agent can either fly over one location or fly across one edge in each point in time (Eq.~\ref{eq:con5}). For each agent and point in time, at most one trip across an edge can start (Eq.~\ref{eq:con7}). If an agent departs from node $n$ at time $t$, then this agent must be traveling across any edge with initial location $n$ at time $t+1$ (Eq.~\ref{eq:con9}) and if an agent arrives at node $n$ at time $t$, then this agent must be traveling across any edge with ending location $n$ at time $t-1$ (Eq.~\ref{eq:con10}). Finally, for each agent and locations, the number of times the $start$ and $k$ decision variables change value form $1$ to $0$ and from $0$ to $1$ must be equal (Eq.~\ref{eq:con12}) and equivalently for each agent and locations, the number of times the $start$ and $ep$ decision variables change  from $1$ to $0$ and from $0$ to $1$ must be equal (Eq.~\ref{eq:con13}). The final two constraints ensure a continuous flying of an agent across a specific pair of nodes. 

The \textit{completion constraints} ensure the proper execution of tasks. Thus, if the initial flying demand for a node is to be covered, at least one agent must fly over this location for a certain time period (Eq.~\ref{eq:con6}). Moreover, for each agent and point in time, if this agent initiates a trip at time $t$, then the starting time of this trip is set to $t$ (Eq.~\ref{eq:con11}). For each agent, the total time it flies across edges must be equal to the travel time required for each trip (Eq.~\ref{eq:con8}). Finally, for each agent, location, and point in time, if a trip begins at time $t$ then this agent must travel for a period of time equal to $t+1+duration$ (Eq.~\ref{eq:con14}).

\section{Greedy scheduling}
\label{sec:greedy}

Given that the optimal solution is practical only for small size problems (see Section~\ref{sec:eval}), here we present a greedy algorithm which applies a one-step look ahead heuristic search mechanism and scales up to problems involving thousands of agents and locations. The algorithm consists of two parts, namely the pre-processing (see Alg.~\ref{alg:schedulingPre1}) and the main scheduling algorithm (see Alg.~\ref{alg:scheduling-random}) parts. 

Regarding the pre-processing part, the initial (Alg.~\ref{alg:schedulingPre1}, lines $1-3$) and final locations (Alg.~\ref{alg:schedulingPre1}, lines $4-6$) of each agent are set. If the start location is different than the final one, the values for the time to travel between these two positions is set to $-1$ (denoted in the greedy algorithm for simplicity as $ep_{a,t} \in L$) (Alg.~\ref{alg:schedulingPre1}, lines $7-8$). Note that these values correspond to the minimum travel time of an agent which is to hover above its initial location until it must start traveling to its final location in order to be there at the last point in time.  

\begin{algorithm}[!htb]
{\small\begin{algorithmic}[1]
\Require $N$ and $A$ and $T$ and $\forall a \in A:$ $n_a^{init}, n_a^{fin}$ and $\forall n \in N$, $\forall t \in T:$ $d_{n,t}$. 
\ForAll{($a \in A$)}
	\ForAll{($t \in T$)} \{Initialize the location of the agents.\}
		\State $ep_{a,t=0}= n_a^{init}$ 
	\EndFor
\EndFor
\ForAll{($a \in A$)}
	\If{$(n_a^{init})\neq n_a^{fin}$} \{Set the value for the final location of the agent.\}
		\State $ep_{a,T-1}= n_a^{fin}$ 
		\ForAll{($t \in T:t\geq |T|-2-c_{\{n_a^{init}, n_a^{fin}\}} \& t< |T|-1$)} \{If the start location is different that the final one, set the values for the time to travel between these two positions to $-1$.\}
			\State {$ep_{a,t}= -1$} 
		\EndFor
	\EndIf
\EndFor

\end{algorithmic}}
\caption{Pre-processing Phase (Initialization of sets and variables).}\label{alg:schedulingPre1}
\end{algorithm}  

In terms of the main scheduling algorithm, we initially create a random sequence $seq$ of all agents. Given this, we iterate through this sequence and for each agent and each point in time we take the current location of the agent (Alg.~\ref{alg:scheduling-random}, line $4-7$). If the agent is not currently traveling across two locations, we check whether a flying demand exists for this location and point in time. If this is the case, the sum of covered demand is increased by one (Alg.~\ref{alg:scheduling-random}, lines $9-12$). In the next step of this algorithm, we need to decide the next location of the agent. To do so, we calculate the distance and demand for all locations, and we schedule the agent to travel to the closest location, including its current one, where demand exists and it has enough time to fly from there to its final destination (Alg.~\ref{alg:scheduling-random}, lines $13-17$). If such a new location is found, then the agent is set to travel to this location for a period of time that corresponds to the actual travel time (Alg.~\ref{alg:scheduling-random}, lines $18-20$), and stay there (Alg.~\ref{alg:scheduling-random}, lines $21-22$) until it is time to fly to its final destination (Alg.~\ref{alg:scheduling-random}, lines $23-25$). The intuition of this heuristic is to avoid having agents being idle for a long period of time, but instead immediately root them to locations with demand. Given that each agent has its own initial location, the performance of this algorithm may be affected by the initial random sequence of the agents. Thus, we execute the algorithm multiple times, each time with a different random sequence for the agents. We continue this execution as long as the solution is improving, or until a number of $threshold$ iterations are made without any improvement on the solution (Alg.~\ref{alg:scheduling-random}, lines $1-2$ and lines $26-29$). Note that trying the entire set of possible sequences of agents would demand $|A|!$ executions of the algorithm. From this point onward, we will refer to this algorithm as \textit{Greedy}.  

\begin{algorithm}[!htb]
\begin{algorithmic}[1]
\Require $N$ and $A$ and $T$ and $\forall a \in A:$ $n_a^{init}, n_a^{fin}$, $\forall n \in N, \forall t \in T:$ $d_{n,t}$ and $threshold$.  
	\State{$count=0, best=0$} \{Initiate auxiliary variables.\}
	\While{($count<threshold$)}
	\State{$totalCov = 0$}
	\State{Create a random sequence $seq$ of the systems' agents.}
		\ForAll{$a \in seq$}\State{Create a random sequence $seq$ of the systems' agents.}
			\ForAll{$t \in T$}
				\State{$curLoc = ep_{a,t}$}
			\\	\{If the drone is not currently traveling between two locations:\}
					\If{($d_{curLoc,t}==1)$} \{If there is demand at the current location and point in time\}
						\State{$totalCov = totalCov + 1$} \{The demand for the current location and point in time is covered.\}
						\State{$d_{curLoc,t}=0$}
					\EndIf \\
					\{Search for the closest location with demand that the drone can reach based on its current location and can then travel in time to its final destination.\}
					\State{$min = 10000$ and $newLoc = -1$}
					\ForAll{$k \in L$}
						\If{($d_{k,t+c_{curLoc,k}+1}==1 \mbox{ \& } t+c_{curLoc,k} + c_{k, n_a^{fin}}<T \mbox{ \& } c_{curLoc,k}< min$)}
							\State{$newLoc = k \mbox{, } min = c_{curLoc,k}$}
						\EndIf
					\EndFor \\ \{If such a location has been found\}
						\ForAll{$t' \in T: t'\geq t+1 \mbox{ and } t'<t+1+c_{curLoc,newLoc}$}
							\State{$ep_{a,t'}=-1$} \{Set the time period that the drone is flying from its current location to the new one.\}
						\EndFor
						\ForAll{$t' \in T: t'\geq t+1+c_{curLoc,newLoc} \mbox{ and } t' <T-2$}
							\State{$ep_{a,t'}=newLoc$} \{Set the time period that the drone is to hover over its new location.\}
						\EndFor
						\If{($t+1+c_{newLoc, n_a^{fin}} == T-2$)} \{If it is time for the drone to fly to its final location.\}
							\ForAll{$t'\in T: t'\geq t+1+c_{newLoc,n_a^{fin}} \mbox{ and } t' <T-2$}
								\State{$ep_{a,t'} = n_a^{fin}$}
							\EndFor
						\EndIf
			\EndFor
		\EndFor
		\If{($best<result$)} \{If the current result is better than the best so far, then update the value for best and set the count equal to zero.\}
			\State{$best=totalCov$, $count =0$}
		\Else
			\State{$count = count + 1$}
		\EndIf
	\EndWhile
\end{algorithmic}
\caption{UAV scheduling algorithm.}\label{alg:scheduling-random}
\end{algorithm}

\section{Performance Evaluation}
\label{sec:eval}

In this section we evaluate our algorithms on a number of settings in order to determine their ability to handle potentially large numbers of locations, points in time, UAVs and demand. In doing so, we consider two settings, one having $7$ locations, $49$ edges (i.e., a fully connected graph), up to $5$ agents, $12$ points in time, and $15\%$ of points in time with flying demand for each location, and a larger one with $20$ locations, $400$ edges, $100$ points in time, up to $15$ agents, and $15\%$ of points in time with flying demand. In all cases the duration of the trip between any two locations is between $1$ and $3$ points in time. The first setting is used to evaluate the scalability of the optimal algorithm and the efficiency of the greedy algorithm compared to the optimal. The second setting is used to evaluate the performance of the greedy algorithm and its ability to handle large numbers of locations, agents, and points in time. The evaluation of our algorithms is executed in three main parts:
\begin{itemize}
	\item EXP1: The execution time and the scalability of the optimal and the greedy algorithm.
	\item EXP2: The performance of the optimal and the greedy algorithm in terms of the average percentage of completed flying demand.
	\item EXP3: The sensitivity of the greedy algorithm on the number of locations, UAVs, points in time, and demand. 
\end{itemize}

\subsection{EXP1: Execution Time and Scalability}
\label{time}
\begin{figure}[!htb]
  \centering
    \includegraphics[width=.85\columnwidth]{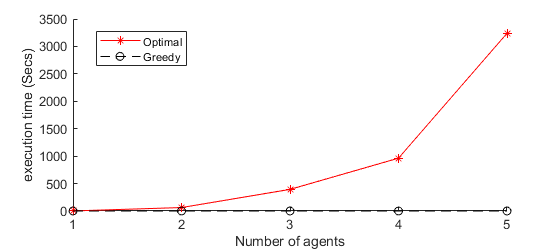}
		\caption{Execution time: Optimal vs greedy algorithm.}
		\label{fig:timeSmall}
\end{figure}

Execution time and scalability are typical metrics for scheduling algorithms. As can be seen from Fig.~\ref{fig:timeSmall}, the execution time of the optimal algorithm shows a steep increase with the number of agents. At the same time, the greedy algorithm has a very low execution time of less than $0.003$ seconds for this small setting which shows a low rate of increase as it remains under $0.5$ seconds even for the larger setting.  

\subsection{EXP2: Performance of the Optimal and the Greedy Algorithms}
\label{perf}
\begin{figure}[!htb]
  \centering
    \includegraphics[width=.85\columnwidth]{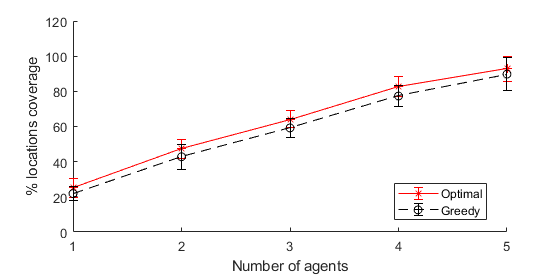}
		\caption{Locations coverage percentage: Optimal vs greedy algorithm.}
		\label{fig:tasksSmall}
\end{figure}

\begin{figure}[!htb]
  \centering
    \includegraphics[width=.85\columnwidth]{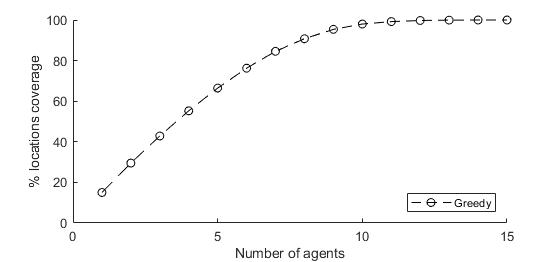}
		\caption{Locations coverage percentage (greedy algorithm).}
		\label{fig:tasksLarge}
\end{figure}

In this section, we evaluate our algorithms in terms of average completion of flying demand. In the small setting and as can be seen from Fig.~\ref{fig:tasksSmall}, the greedy algorithm performs well with performance reaching $96.81\%$ of the optimal in the best case, $87.04\%$ in the worst case, and $92.06\%$ on average. In the large setting and as can be seen from Fig.~\ref{fig:tasksLarge}, the greedy algorithm achieves an approximately $100\%$ completion of flying demand when $15$ drones are used. However, even with $8$ drones, the coverage is already over $90\%$, and with $11$ drones it is over $99\%$. Given the limited scalability of the optimal algorithm, we argue that the performance of the greedy algorithm is very satisfactory, making it the best choice for medium- and large-scale settings. 

\begin{figure}[!htb]
  \centering
    \includegraphics[width=.85\columnwidth]{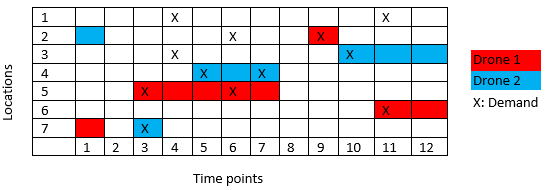}
		\caption{Example execution of the greedy algorithm.}
		\label{fig:example}
\end{figure}

Figure~\ref{fig:example} depicts an example execution of the greedy algorithm for the small setting and for two drones. At times $1$ and $12$, the drones are at their initial and final locations respectively. At the rest of the points in time, the algorithm tries to cover the demand to the maximum extend. In doing so, some of the demand is impossible to be covered as for example at point in time $3$ at which time both drones are covering demand at locations $5$ and $7$, and so by definition they cannot be at any other location at exactly the next point in time. The same is true for locations $2$ and $1$ and times $6$ and $11$, respectively. 
Note that the cells which are not colored are the ones where the drones are flying across two locations.

\subsection{EXP3: Sensitivity of the Greedy Algorithm}
\label{sens}
\begin{figure}[!htb]
  \centering
    \includegraphics[width=0.85\columnwidth]{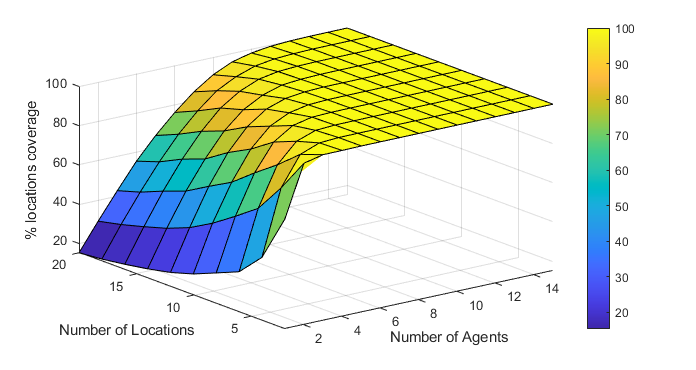}
		\caption{Sensitivity of the greedy algorithms when varying the number of agents and locations.}
		\label{fig:SurfLocsAgents}
\end{figure}

The greedy algorithm has already shown to have near optimal performance. However, we also need to evaluate how this performance may be affected by the number of locations, points in time, and density of demand. As can be seen from Fig.~\ref{fig:SurfLocsAgents} when the number of locations increases, the performance of the algorithm gradually drops and close to $100\%$ coverage is achieved only with high number of drones. This occurs because when the locations increase, the total demand also increases. Further, when the locations increase, the drones tend to fly around more as the demand is more spatially scattered. 

\begin{figure}[!htb]
  \centering
    \includegraphics[width=0.85\columnwidth]{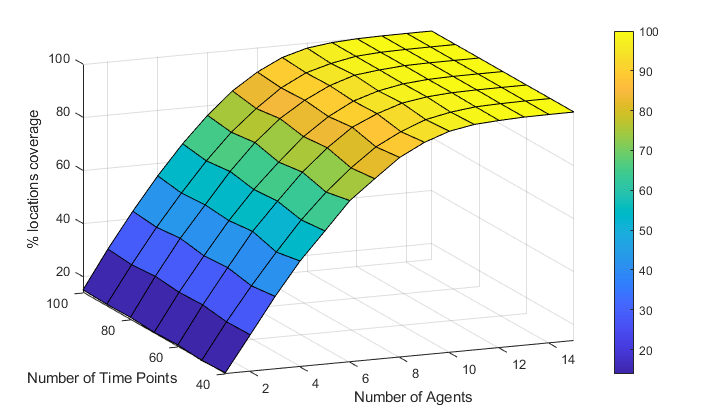}
		\caption{Sensitivity of the greedy algorithm when varying the number of agents and points in time.}
		\label{fig:SurfTimePointsAgents}
\end{figure}

In addition, and as can be seen from Fig.~\ref{fig:SurfTimePointsAgents}, when the locations remain fixed, but the number of the points in time increases we observe no major change in the performance of the algorithm. It is important to note that in this case the volume of points in time with demand increases, but the percentage of the total remains fixed at $15\%$. Thus, from these two experiments we can conclude that the performance of the algorithm is affected by the number of locations, but not from the number of points in time. 

\begin{figure}[!htb]
  \centering
    \includegraphics[width=0.85\columnwidth]{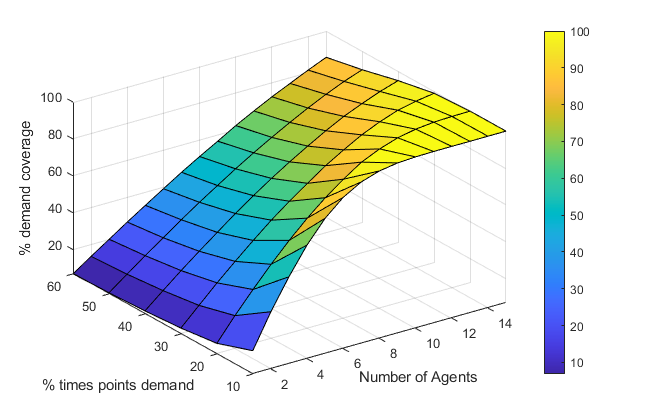}
		\caption{Sensitivity of the greedy algorithms when varying the number of agents and points in time with demand.}
		\label{fig:SurfAgentsDemand}
\end{figure}

Another dimension that has to be examined, is how the greedy algorithm is affected by the density of the demand, in other words the percentage of points in time with fly over demand. As can be seen from Fig.~\ref{fig:SurfAgentsDemand}, when the density of the demand is high, the performance of the algorithm drops. For instance, if we have $15$ drones and $15\%$ demand the coverage is at approximately $100\%$, but when the demand is at $60\%$ the coverage is at approximately $85\%$. To explain our decision to experiment with up to $60\%$ demand, we argue that if the demand is too high, this leads to a situation where a drone should constantly fly over each location. In such a setting, a scheduling algorithm is not really needed. 

\section{Conclusions and future work}
\label{sec:con}

In this work we examined the scheduling of drones across a graph. In this vein we extended the well-known multiple-traveling salesman problem by adding the constraint of multiple visits per node at specific points in time. Initially, we formulated the problem as an Integer Linear Program and we solved it offline and optimally. Given that this solution has limited scalability, we also developed a greedy algorithm that uses a one-step look-ahead heuristic function and achieves near optimal performance while also scaling to large settings.  

For future work, we aim to handle the limited range of the drones by adding the ability to recharge their batteries between specific routes. We also aim to monitor and manage the flying altitude of the drones to achieve collision avoidance. Finally, we aim to develop an online algorithm for the same problem that will use reinforcement learning techniques. 

\section*{ACKNOWLEDGMENT}
This work was supported by the European Union's Horizon 2020 research and innovation programme under grant agreement No 739551 (KIOS CoE) and from the Government of the Republic of Cyprus through the Directorate General for European Programmes, Coordination and Development.

\bibliographystyle{IEEEtran}
\bibliography{literature_GEv2}

\end{document}